\documentclass{article} 
\usepackage[preprint]{colm2025_conference}

\usepackage{microtype}
\usepackage{hyperref}
\usepackage{url}
\usepackage{booktabs}
\usepackage{multirow}
\usepackage{graphicx}
\usepackage{paralist}
\usepackage{longtable}

\usepackage{lineno}
\usepackage{cleveref}

\definecolor{darkblue}{rgb}{0, 0, 0.5}
\hypersetup{colorlinks=true, citecolor=darkblue, linkcolor=darkblue, urlcolor=darkblue}

\makeatletter
\def\@fnsymbol#1{\ensuremath{\ifcase#1\or \dagger\or \ddagger\or
   \mathsection\or \mathparagraph\or \|\or **\or \dagger\dagger
   \or \ddagger\ddagger \else\@ctrerr\fi}}
\makeatother
    
\title{\textsc{LogicPO}: Efficient Translation of NL-based Logical Problems to FOL using LLMs and Preference Optimization}

\author{Koushik Vishwanadha$^1$\thanks{Work done as part of an internship under Tr$^2$AIL Lab, IIT Kharagpur}, Deepanway Ghosal$^2$$^\dagger$ \& Somak Aditya$^3$ \\
$^1$Indeed Inc. \\
$^2$Singapore University of Technology and Design \\
$^3$IIT Kharagpur\\
\texttt{koushik.viswanadha@alumni.iiit.ac.in, saditya@cse.iitkgp.ac.in} 
}

\begin{document}

\ifcolmsubmission
\linenumbers
\fi

\maketitle

\begin{abstract}
Logical reasoning is a key task for artificial intelligence due to it's role in major downstream tasks such as Question Answering, Summarization. Recent neurosymbolic methods in improving the reasoning ability of Large Language Models (LLM) fall short in correctly converting a natural language reasoning problem to an equivalent logical formulation, which hinders the framework's overall ability to reason. Towards this, we propose to use finetuning on a preference optimization dataset to learn to translate a natural language reasoning problem \textit{in its entirety} to a consistent logical program by 1) introducing a new supervised and preference optimization dataset (\textsc{LogicPO}), and 2) adopting popular techniques such as Direct Preference Optimization (DPO), Kahneman-Tversky optimization (KTO) to finetune open-source LLMs. 
Our best model with \textsc{Qwen-2.5} (14B) consistently outperforms \textsc{GPT-4}'s (8-shot) by producing $6\%$ more logically correct and with $8\%$ less syntax errors. We show that translating problems as a whole significantly surpasses sentence-wise text to First order Logic (FOL) baselines. We further explicitly discuss the categories of errors that our framework addresses (and does not address), in the context of recent comparable Neurosymbolic provers\footnote{The code and data is available at \url{github.com/goku80903/LogicPO}.}.  
\end{abstract}

\section{Introduction}


Recent state-of-the-art pipelines for logical reasoning tasks show a marked shift from having Large Language Models (LLMs) generate solutions directly. Most frameworks first translate the problem into a formal language using an LLM and then use a corresponding logical engine (such as a theorem prover) to solve reasoning tasks \citep{logiclmemnlpPanAWW23,satlmnipsYeCDD23,lincOlaussonGLZSTL23}.  
For example, Logic-LM \citep{logiclmemnlpPanAWW23} demonstrates how tasks can be converted into equivalent first-order logic, satisfiability, or constraint satisfaction problems, and solved using publicly available symbolic engines.
However, the efficiency of these methods is limited as pretrained LLMs make various syntactic, logical and semantic mistakes while translating the reasoning problem from text to a \textit{consistent} logical language. This is possibly because, these logical languages are \textit{low-resource} compared to coding languages such as Python, and therefore less prevalent in pretraining data.

In this work, we therefore focus solely on analyzing and then improving the translation quality of a natural language-based reasoning \textit{problem}  (including a \textit{context} and a \textit{query}) consistently to First Order Logic (FOL) problem. We choose FOL as the target language, as it is a widely adopted logical language with broad expressivity, and one can utilize available algorithms to convert FOL programs to other equivalent languages (SAT, SMT).  We employ three logical reasoning datasets (with available FOL translations per problem) for benchmarking. 
We observe that contemporary efforts to improve FOL translations primarily target sentence-wise translation \citep{yang-etal-2024-harnessing}, which does not take into account predicate level consistency -- for example, \textit{Tab.~\ref{tab:folio-example} shows how the last sentence is translated into \texttt{SAT2016(.)} predicate, which does not match \texttt{SAT(.)} predicate used elsewhere.}

Large Language Models (such as \textsc{GPT3.5-Turbo}, Llama-3-8B), even with in-context learning, makes similar mistakes alongwith various syntactic, logical and semantic errors. 
We therefore resort to finetuning open-source LLMs for the task of problem-level text-to-FOL translation. We start with a seed dataset called FOLIO \citep{han-etal-2024-folio}, where each example consists of a natural language problem (story $S$ having context and query), a logical label (\textit{True}, \textit{False} or \textit{Uncertain}) and parallel FOL story ($FS_{gt}$). We use the natural language stories from FOLIO as input, and perform in-context learning with a few groundtruth NL-FOL paired samples on various state-of-the art LLMs to generate potential FOL stories. For each input story, we vary the number of in-context examples, and temperature to generate multiple candidate FOL stories ($\tilde{FS}$). We then use a logical prover (Prover9\footnote{\url{https://www.cs.unm.edu/~mccune/prover9/}}) to obtain the logical label from each generated FOL story ($\tilde{FS}$). If the computed logical outcome matches with the original groundtruth label, we include the NL-FOL pair ($S,\tilde{FS}$) as part of our supervised finetuning ($\mathcal{D}_{sft}$) dataset. For our preference dataset ($\mathcal{D}_{pref}$), for each input story ($S$), we include two FOL stories -- one with \textit{correct} logical label (same as groundtruth), and one with \textit{incorrect} logical label (or resulting in errors from Prover9).
We first finetune the open-source LLMs on $\mathcal{D}_{sft}$ using supervised finetuning. Then, we adopt various preference optimization methods (DPO and KTO) and use them to further finetune the large language models on $\mathcal{D}_{pref}$. We thoroughly analyze the performance of these finetuned models with respect to a large set of baseline LLMs (while observing the effect of increasing in-context examples). Lastly, we analyze what errors remain difficult to reduce even after employing our dataset and training methodologies. In summary, our contributions are three-fold:
\begin{compactitem} 
    \item We introduce a novel way to bootstrap a large supervised finetuning and a preference optimization dataset (\textsc{LogicPO} with $26k$) for natural language reasoning problem to FOL translation, starting from a small seed dataset (FOLIO with $1k$). Here, the input is textual context and query, and the output is a correct (or incorrect) FOL program, capturing various (implicit) perturbations to curb errors at a logical, syntactic, and semantic level.
    \item We show the utility of our dataset by employing various preference optimization methods (DPO and KTO) on open-source LLMs such as Llama-3, Gemma-3, and Phi-3.5. We observe significant decrease in syntactic and logical errors -- indicating successful end-to-end solution of the input reasoning problem. Our finetuned models (especially Llama-3-8B and Phi-3.5-Mini with KTO) surpass all sentence-wise baselines, GPT3.5 variants, and rank close to the baseline of GPT4 (8 shots) with majority voting strategies \citep{lincOlaussonGLZSTL23}.
    \item Lastly, we discuss the strengths and shortcomings of the \textsc{LogicPO}-finetuned models in the context of contemporary neurosymbolic provers. We show that while our methodology decreases the use of inconsistent predicates and improves on syntactic errors, our finetuned models suffer from information leakage -- leaving room for future improvements.
\end{compactitem}

\section{Related Work}
In NLP, prior to Transformers-based end-to-end systems, the need for Semantic Parsers dominated various downstream applications, such as Question-Answering, and Recognizing Textual Entailment \citep{Gu2022KnowledgeBQ,beltagy-etal-2014-probabilistic,DBLP:conf/ijcai/DengZCWR22}. The task of learning to parse involved: 1) converting natural language to expert-defined novel semantic representations \citep{banarescu-etal-2013-abstract, DBLP:conf/conll/GeM05,DBLP:conf/ijcai/SharmaVAB15,chanin2023opensource} or 2) translating text to sentences in well-established programming languages such as SQL, and First Order Logic. 
Several solutions around knowledge-based question answering \citep{DBLP:journals/pvldb/KimSHL20,DBLP:conf/cvpr/WuWSDH16,DBLP:journals/corr/abs-2408-05109} attempted to convert the natural language question to structured SQL queries (or SPARQL queries) against known schemas. Such models demanded a large number of parallel data points (source sentence and target parse), and did not explore context-level inter-sentence consistency.


In recent times, various researchers have cast doubt on whether LLMs actually reason \citep{nikankin2024arithmeticalgorithmslanguagemodels,DBLP:conf/acl/LuBSMG24}. As a remedy, a series of efforts \citep{lincOlaussonGLZSTL23, logiclmemnlpPanAWW23,satlmnipsYeCDD23} have shown that LLMs can be used as few-shot translators to represent the reasoning problem into an equivalent programming language; and such programs can be executed to find the (correct) answer to the problem. LLMs have shown to excel at translating natural language reasoning problems into Python via in-context learning but struggle with less common formal languages like First-order Logic. Researchers resort to multiple heuristics, such as self-refinement \citep{lincOlaussonGLZSTL23} using error(s) signals from Automated Theorem Provers (such as Z3, Prover9) as additional input. These are quite cost-extensive with multiple LLM calls, and do not greatly improve the translation accuracy. There have been some efforts using small Language Models (such as T5) \citep{lu-etal-2022-parsing} to learn to parse Natural Language sentences to FOL, most efforts concentrate on benchmarking sentence level parsing, such as \textsc{LogicLLAMA} \citep{yang-etal-2024-harnessing}. The authors in \textsc{LogicLLAMA} utilize GPT-4 sentence-level NL-FOL pairs, followed by filtering to create a data set and use supervised fine-tuning to finetune a Llama model. A recent non-peer reviewed paper introduces ProofFOL \citep{thatikonda2024strategies}, a high-quality FOL-annotated subset of ProofWriter dataset using GPT-4o. While this dataset is partly useful, the dataset and the trained models are not publicly available. 

In contrast, our goal is to convert a natural language reasoning problem as a whole to a consistent FOL program. We introduce a preference dataset, perform an extensive study utilzing supervised finetuning and many preference optimization techniques on multiple open-source LLMs. 
To the best of our knowledge, our work is the first to explore Preference Optimization techniques to train open-source LLMs (such as Llama) to reduce synactic, and logical errors in FOL parsing. We also plan to release our preference dataset (\textsc{LogicPO}), that can enable new algorithms and new family of FOL translators.


\section{Learning Preference-Optimized NL to FOL Engines}






\subsection{Natural Language to First-Order Logic Conversion}
\label{FOLIO-explanation}
The FOLIO dataset \citep{han2024folionaturallanguagereasoning} is an expert-written dataset containing high-quality examples requiring complex logical reasoning in FOL. The dataset consists of two tasks: natural language reasoning with FOL and NL to FOL translation. We are interested in the second task of NL to FOL translation in this work. The goal of this task is to translate an NL story $S$ to a FOL story $FS$. The NL story $S$ contains a series of context sentences $p_1, p_2, \dots p_n$ and a conclusion sentence $p_{n+1}$. The FOL story $FS$ consists of context formulas $f_1, f_2, \dots f_n$ and a conclusion sentence $f_{n+1}$. The translation task requires each $p_i$ in $S$ to be logically and semantically equivalent to its corresponding FOL formula $f_i$ in $FS$. Moreover, the logical values for the conclusions $p_{n+1}$ and $f_{n+1}$ should be the same based on the corresponding context $S$ and $FS$. We show an example of the task in Table~\ref{tab:folio-example}.

\begin{table*}[ht]
\small
\centering
\resizebox{0.85\textwidth}{!}{
\begin{tabular}{@{}l|l@{}}
\toprule[1pt]
\begin{tabular}[c]{@{}l@{}}FOLIO\\  Instance\end{tabular}                      & \begin{tabular}[c]{@{}l@{}}
\textbf{Premises}: \\ 1. The SAT test is wholly owned and developed by the College Board.\\ 2. The SAT test is intended to assess student's readiness for college. \\ 3. The SAT was originally designed not to be aligned with high school curricula.\\ 4. Several adjustments were made for the version of the SAT introduced in 2016 \\to reflect more closely on what students learn in high school.\\ 
\textbf{Premises-FOL}: \\ 
1. $Own(sat, collegeBoard) \land \neg Own(sat, others)$ \\ 
2. $Test(sat, readiness)$ \\ 
3. $\forall x (Year(x) \land Before2016(x) \Rightarrow \neg AlignHighSchool(x))$ \\ 4. $\exists x (Year(x) \land Since2016(x) \land AlignHighSchool(x))$\\ 
\textbf{Conclusion}: Since 2016 the SAT has been better aligned with high school curricula.\\ 
\textbf{Label}: [\textit{True}]\end{tabular} \\ \midrule

\begin{tabular}[c]{@{}l@{}}$\mathcal{D}_{sft}$ Output, \\ $\mathcal{D}_{pref}$ Chosen\end{tabular} & \begin{tabular}[c]{@{}l@{}}\textbf{Premises-FOL}: \\ 
1. $\forall x. (SAT(x) \Rightarrow CollegeBoardOwns(x))$\\
2. $\forall x. (SAT(x) \Rightarrow CollegeReady(x))$\\ 
3. $\forall x. (SAT(x) \land \neg Aligned(x))$\\ 
4. $\forall x. (SAT(x) \land IntroducedIn2016(x) \Rightarrow Aligned(x))$\\ 
\textbf{Conclusion-FOL}:  $\forall x. (IntroducedIn2016(x) \Rightarrow Aligned(x))$\\ 
\textbf{Label}: [\textit{True}]\end{tabular}
\\ \midrule

\begin{tabular}[c]{@{}l@{}}$\mathcal{D}_{pref}$ Rejected\end{tabular}  & \begin{tabular}[c]{@{}l@{}}\textbf{Premises-FOL}:\\ 
1. $\forall x. (SAT(x) \Rightarrow CollegeBoard(x))$\\ 
2. $\forall x. (SAT(x) \Rightarrow CollegeReady(x))$
\\ 3. $\forall x. (SAT(x) \land \neg AlignedWithHighSchool(x))$
\\ 4. $\forall x. (SAT2016(x) \Rightarrow AlignedWithHighSchool(x))$\\ 
\textbf{Conclusion-FOL}: $\forall x. (Since2016(x) \land AlignedWithHighSchool(x))$\\ \textbf{Label}: [\textit{False}]\end{tabular}\\ \midrule[1pt]


\end{tabular}
}
\caption{\small{An example from the FOLIO dataset and its corresponding versions in our $\mathcal{D}_{sft}$ and $\mathcal{D}_{pref}$ datasets. The Label information in $\mathcal{D}_{sft}$ and $\mathcal{D}_{pref}$ is shown for completeness. It is not used during training or inference of our models.}}
\label{tab:folio-example}
\end{table*}

\begin{figure*}[!ht]
    \centering
    \includegraphics[width=\textwidth]
    {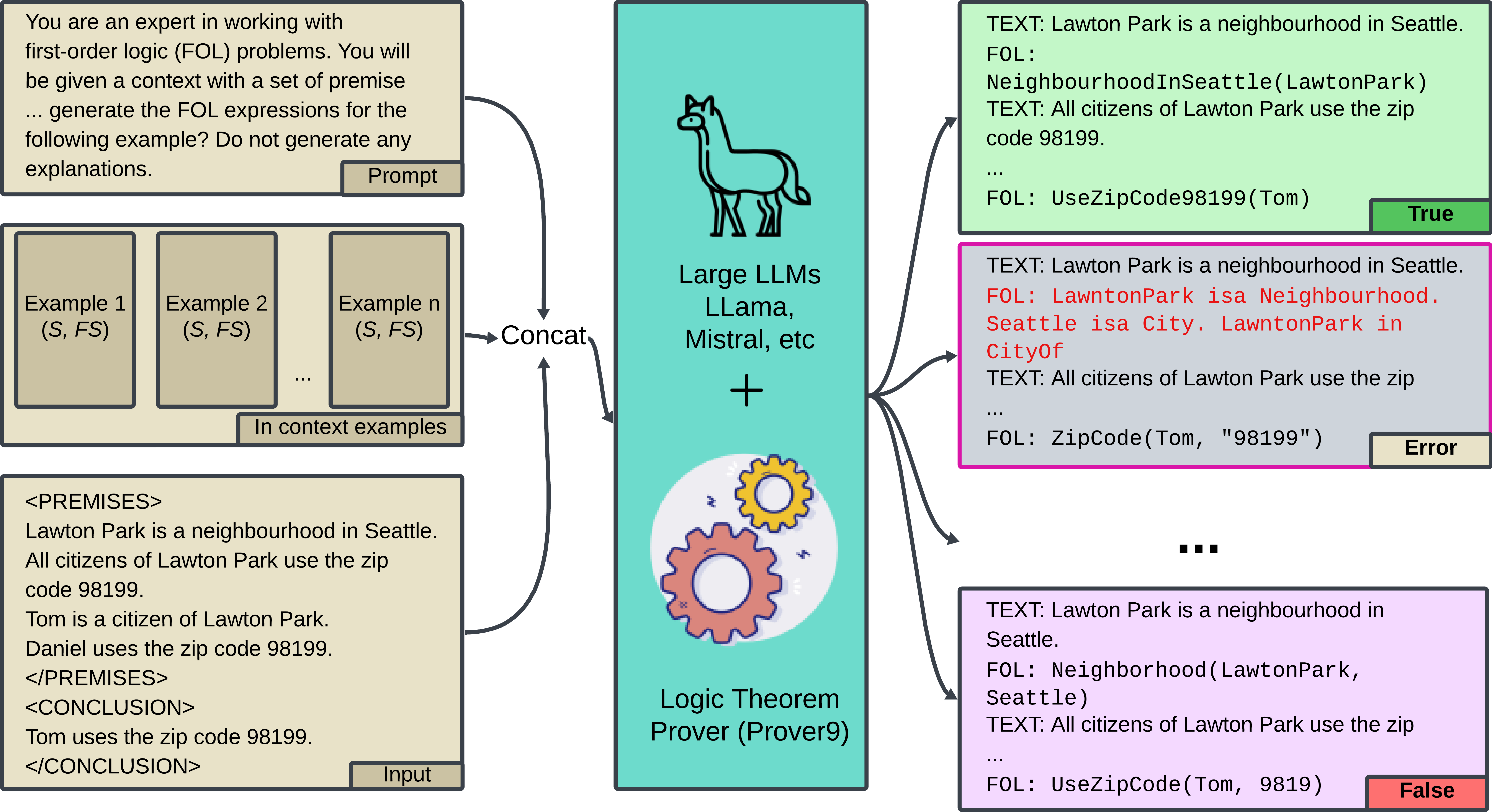}
    \caption{The Data creation pipeline: We use natural language stories from FOLIO, use different settings of LLMs to create various first order logic parses. We then use an off-the-shelf theorem prover (such as Prover9) to observe the predicted logical label of the conclusion. If the predicted label matches the original, we include the sample in $\mathcal{D}_{sft}, \mathcal{D}_{pref}$ (chosen) and if it does not match, we include it as $\mathcal{D}_{pref}$ rejected sample.}
    \label{fig:enter-label}
\end{figure*}

\subsection{Automated Data Generation}
\label{sec:data-gen}
Let's denote $X$ to be an instance of NL to FOL conversion data consisting of an NL story $S$ and a FOL story $FS$. Let's also consider $y$ to be the logical label of the conclusion sentence/formula constrained on the contextual sentences/formulas. The logical label $y$ always belongs to the set of \{\textit{True, False, Uncertain}\}. 

\paragraph{Setup}
We first follow the following pipeline to generate a collection of output FOL stories $\bar{FS}$ from given input NL stories $S$:

\begin{compactitem}
    \item Consider $n$ different examples of $X$ from FOLIO to use as in-context demonstrations.
    \item Define an appropriate natural language prompted instruction for NL to FOL conversion. Additionally use the $n$ demonstrations in context and ask an LLM to generate the output FOL story $\bar{FS}$ for an input NL story $S$.
    \item Consider the logical label of $S$ to be $y_{nl}$. Predict the logical label $\bar{y}_{fs}$ of the generated story $\bar{FS}$ using a standalone logical engine such as Prover9 \citep{prover9}. Note that the generated logical label $\bar{y}_{fs}$ could be either from \{\textit{True, False, Uncertain}\} or could also be a \{\textit{Error}\} in case the generated $\bar{FS}$ cannot compile in the logical engine for any reason.
\end{compactitem}

\paragraph{Dataset for Supervised Fine-Tuning.}
We construct the supervised fine-tuning (SFT) data by considering instances for which the generated FOL logical label $\bar{y}_{fol}$ matches the given natural language logical label $y_{nl}$. These SFT instances consist of $S$ as the input and $\bar{FS}$ as the output, as we can consider $\bar{FS}$ to be an equivalent version of $S$ because they both lead to the same logical label. 

\paragraph{Dataset for Preference Optimization.}
Creating preference optimization data requires a preferred and a rejected output sample for the same input. We consider the generated $\bar{FS}$ for which the FOL logical label $\bar{y}_{fol}$ matches the natural language logical label $y_{nl}$ as the preferred sample. On the other hand, $\bar{FS}$ for which the label $\bar{y}_{fol}$ does not match $y_{nl}$ can be considered as the rejected sample. In this case, the preferred and the rejected FOL samples can be considered as the clean and noisy translations of the natural language story, respectively.

\paragraph{Summary of Generated Datasets}
We denote the SFT dataset as $\mathcal{D}_{sft}$ and the preference dataset as $\mathcal{D}_{pref}$. The summary of the datasets is shown in \Cref{tab:data-stat}.

The input NL stories are drawn from the training set of FOLIO. We used $n = 2, 4, 8$ in-context examples to generate the output FOL stories, as mentioned earlier in ~\ref{sec:data-gen}. We used the instruction-tuned 8B versions of Llama 3 \citep{grattafiori2024llama3herdmodels}, Llama 3.1 \citep{grattafiori2024llama3herdmodels}, and the 7B version of Mistral 0.2 \citep{mistral7bjiangetal23} as the language models to generate the samples at temperatures of $0.25$ and $0.6$. We follow this generation setup with multiple LLMs at different temperatures with different in-context LLMs to increase the diversity of the generated outputs. In total, we generate $30$ output samples for each input NL story in FOLIO. From there, we randomly subsample and select outputs $\bar{FS}$ to create $\mathcal{D}_{sft}$ and $\mathcal{D}_{pref}$, following the same/different logical label matching strategy, as specified in section~\ref{sec:data-gen}. 

\begin{table}[t!]
\small
\centering
\resizebox{0.45\linewidth}{!}{
\begin{tabular}{l|ccc}
\toprule[1pt]
Logical Label & FOLIO & $\mathcal{D}_{sft}$ & $\mathcal{D}_{pref}$ \\
\midrule
True & 388 & 5,001 & 3,751 \\
False & 286 & 3,169 & 2,600 \\
Uncertain & 330 & 8,792 & 3,649 \\
\midrule
Total & 1,004 & 16,962 & 10,000\\
\bottomrule
\end{tabular}
}
\caption{\small{Number of instances in FOLIO, $\mathcal{D}_{sft}$ and $\mathcal{D}_{pref}$ across the different logical labels. The logical label correspond to the label of the preferred sample for $\mathcal{D}_{pref}$.}}
\label{tab:data-stat}
\end{table}

\subsection{Language Models as NL to FOL Engines}
Our objective is to design language models that can convert an NL story $S$ to its corresponding logically consistent\footnote{We use \textit{consistency} to imply inter-sentence predicate-level consistency, so that a logical system (such as Z3, Prover9) can carry out resolution refutation-style algorithms.} FOL story $FS$. We hypothesize that our automatically generated datasets $\mathcal{D}_{sft}$ and $\mathcal{D}_{pref}$ would be useful in teaching language models to become effective NL to FOL conversion engines.

Firstly, $\mathcal{D}_{sft}$ provides parallel NL to FOL conversion data over all the three logical labels of \{\textit{True, False, Uncertain}\}. This signal would help language models to observe diverse stories with conclusive and inconclusive scenarios. Secondly, $\mathcal{D}_{pref}$ provides signals of what is and isn't the correct FOL conversion of an NL story. The additional signal of how not to convert NL to FOL stories comes from the rejected samples. This would help in teaching language models to avoid issues like syntax errors, which are common in non-fine-tuned models. The rejected samples also work as hard negatives, as the corresponding FOL story is still closely related to the NL story while being logically inconsistent. 

We follow the strategy of i) fine-tuning the language model with $\mathcal{D}_{sft}$ and then ii) fine-tuning with $\mathcal{D}_{pref}$ using commonly used preference optimization algorithms such as DPO and KTO \citep{ethayarajh2024ktomodelalignmentprospect}. We fine-tune all the parameters of the language model in each of the two stages. We empirically show that this is a useful strategy in making language models highly effective NL to FOL conversion engines. We also analyze the results across various dimensions to find interesting insights (\S\ref{sec:experiments}).

\section{Experiments}
\label{sec:experiments}
In this section, we present our experimental setup, datasets, and the baseline models used to compare our parsers. We also give a detailed study of the current parsing abilities of various SOTA LLMs.

\subsection{Datasets}
Our experiments use tasks from three existing datasets: \textbf{FOLIO}~\citep{han2024folionaturallanguagereasoning},  \textbf{ProofWriter}~\citep{tafjord2021proofwritergeneratingimplicationsproofs} and \textbf{PrOntoQA}~\citep{PrOntoQA}, all of which have been shown to be challenging for off-the-shelf LLMs \citep{lincOlaussonGLZSTL23}\footnote{We further provide inference results on recently published ProverQA \citep{qi2025largeproverqa} in \S\ref{sec:app:addl:proverqa}.}. The FOLIO dataset is an expert-written dataset containing high-quality examples requiring complex logical reasoning in FOL. As mentioned in section ~\ref{FOLIO-explanation}, we focus on the NL to FOL translation task. We use its validation set for our evaluation, containing 204 examples. Meanwhile, ProofWriter is a synthetically generated dataset for logical reasoning over natural language. We use the sampled split of 360 data points provided by \citet{lincOlaussonGLZSTL23} for our experiments. The data points are uniformly distributed across the three labels (\textit{True, False, Uncertain}) and maximum question depth (ranging from 0-5; 50 samples each). ProntoQA is another synthetically generated question-answering dataset, converted from a synthetic world model represented in First Order Logic. It contains triplets of context, query, and a label. We use the validation split that has 500 instances across the (\textit{True, False}) labels. We change the objective of the dataset from question answering to NL-FOL conversion task.

\paragraph{Metrics.} We follow \cite{lincOlaussonGLZSTL23} to report logical correctness, incorrectness and syntax errors. Deviating from the k-majority voting practice, for each NL story, we generate 10 outputs. For each output FOL story, the Prover9 provides a predicted logical label (or generates syntax error). For correctness, we report the average number of matches to the original label. For incorrectness, we report the average number of times the output label is incorrect. Similar goes for syntax error, the average number of times the output is syntactically incorrect. We additionally report the overall weighted F1 and the F1 score over the True labels.

\begin{table*}[!ht]
\small
\centering
\resizebox{0.9\textwidth}{!}{
\begin{tabular}{@{}ll|ccc|cc@{}}
\toprule
\multirow{2}{*}{\textbf{Model}} & \multirow{2}{*}{\textbf{Setting}} & \multicolumn{2}{c}{\textbf{Logically}} & \textbf{Syntax} & \multicolumn{2}{c}{\textbf{F1}} \\ 
& & \textbf{Correct (↑)} & \textbf{Incorrect} & \textbf{Error (↓)} & \textbf{Overall (↑)} & \textbf{True Label (↑)}\\
\midrule

\multirow{4}{*}{GPT-3.5 - \textsc{LINC}} & 1-shot & 33.30 & 15.44 & 51.26 & 40.49 & 22.71 \\
     & 2-shot & 45.82 & 23.24 & 30.93 & 38.67 & 52.32 \\
     & 4-shot & 51.81 & 24.45 & 23.74 & 45.50 & 59.10 \\
     & 8-shot & 51.31 & 24.78 & 23.90 & 45.63 & 53.29 \\ 
\midrule
\multirow{2}{*}{GPT-3.5 - Sentence-wise} & 4-shot sentence & 6.86 & 8.83 & 84.31 & 10.17 & 5.33  \\
     & 4-shot paragraph & 21.54 & 35.39 & 43.07 & 23.96  & 29.27 \\
\midrule
\multirow{1}{*}{LogicLLama - Sentence-wise} & 5-shot sentence & 16.67 & 22.55 & 60.78 & 17.88 & 5.33  \\
\midrule
\multirow{1}{*}{GPT-4 - \textsc{LINC}} & 8-shot & 64.01 & 23.08 & 12.91 & 59.89 & 67.00  \\
\midrule
\multirow{4}{*}{Llama-3 8B Instruct} 
     & 2-Shot & 50.15 & 27.99 & 21.86 & 55.43 & 54.11\\
     & SFT & 54.26 & 34.26 & 11.47 & 56.73 & 58.84 \\
     & SFT + DPO & 50.98 & 29.51 & 19.51 & 55.80 & 56.47\\
     & SFT + KTO & 55.15 & 30.98 & 13.87 & 59.56 & 61.74\\
\midrule
\multirow{3}{*}{Gemma-2 2B Instruct} 
& 2-Shot & 10.29 & 40.21 & 49.50 & 14.50 & 28.57\\
     & SFT & 49.71 & 34.80 & 15.49 & 53.12 & 54.51 \\
     & SFT + DPO & 34.95 & 21.23 & 43.82 & 44.09 & 42.30 \\
     & SFT + KTO & 50.93 & 29.80 & 19.26 & 56.02 & 54.39\\
\midrule
\multirow{3}{*}{Phi-3.5 Mini Instruct (4B)} 
     & 2-Shot & 6.37 & 15.2 & 78.43 & 11.18 & 16.47 \\
     & SFT & 58.43 & 31.08 & 10.49 & 61.47 & 62.03\\
     & SFT + DPO & 61.13 & 31.08 & \textbf{7.79} & 63.59 & 60.73\\
     & SFT + KTO & \textbf{61.52} & 29.41 & 9.07 & \textbf{64.43} & \textbf{63.76}\\\midrule
\multirow{3}{*}{Qwen-2.5 Instruct (14B)} 
     & 2-Shot & 60.78 & 29.42 & 9.80 & 63.44 & 61.67 \\
     & SFT       & 57.84             & 28.97               & 13.19        & 61.10      & 60.17         \\ 
& SFT + DPO & 68.82             & 26.62               & 4.56         & 70.36      & 72.03         \\
& SFT + KTO & \textbf{70.20 }            & \textbf{25.25 }              & \textbf{4.56}         & \textbf{71.78}      & \textbf{73.58}         \\
\bottomrule
\end{tabular}}
\caption{Results on NL to FOL conversion on the FOLIO validation dataset. The results are an average of 10 runs (5 in case of few-shot). The base prompts are in Appendix.}
\label{tab:results_FOLIO}
\end{table*}

\begin{table*}[!ht]
\small
\centering
\resizebox{0.9\textwidth}{!}{
\begin{tabular}{@{}ll|ccc|cc@{}}
\toprule
\multirow{2}{*}{\textbf{Model}} & \multirow{2}{*}{\textbf{Setting}} & \multicolumn{2}{c}{\textbf{Logically}} & \textbf{Syntax} & \multicolumn{2}{c}{\textbf{F1}} \\ 
& & \textbf{Correct (↑)} & \textbf{Incorrect} & \textbf{Error (↓)} & \textbf{Overall (↑)} & \textbf{True Label (↑)}\\
\midrule

\multirow{3}{*}{Llama-3 8B Instruct} 
     & 2-shot & 59.17 & 28.05 & 12.78 & 63.15 & 62.45 \\
     & 4-shot & 25.83 & 51.94 & 22.22 & 15.50 & 46.50 \\
     & SFT & 46.14 & 42.39 & 11.47 & 48.04 & 38.80 \\
     & SFT + DPO & 45.31 & 35.47 & 19.22 & 50.16 & 45.55 \\
     & SFT + KTO & 48.58 & 38.06 & 13.36 & 51.58 & 44.12 \\
\midrule
\multirow{3}{*}{Gemma-2 2B Instruct} 
     & 2-shot & 10.56 & 17.77 & 71.67 & 14.84 & 17.72 \\
     & 4-shot & 11.11 & 23.06 & 65.83 & 10.97 & 32.92 \\
     & SFT & 46.78 & 32.56 & 20.67 & 51.74 & 45.91 \\
     & SFT + DPO & 39.97 & 22.50 & 37.53 & 48.69 & 42.12 \\
     & SFT + KTO & 48.86 & 32.94 & 18.19 & 53.65 & 49.47 \\
\midrule
\multirow{3}{*}{Phi-3.5 Mini Instruct (4B)} 
     & 2-shot & 30.56 & 5.1 & 64.44 & 44.74 & 42.04 \\
     & 4-shot & 18.89 & 5.83 & 75.28 & 29.88 & 22.70 \\
     & SFT & 61.11 & 30.0 & 8.89 & 63.49 & 54.22 \\
     & SFT + DPO & 55.00 & 36.00 & 9.00 & 56.44 & 46.04 \\
     & SFT + KTO & \textbf{65.28} & 27.89 & \textbf{6.83} & \textbf{67.43} & \textbf{58.96} \\
\midrule
\multirow{3}{*}{Qwen-2.5 Instruct (14B)} 
     & 2-Shot & 87.22 & 10.28 & 2.5 & 88.39 & 85.58 \\
     & 4-Shot & 89.72 & 8.88 & 1.4 & 90.34 & 86.64 \\
     & SFT       & 88.89    & 9.91       & 1.20 & 89.45 & 85.99 \\ 
& SFT + DPO & 88.80    & 9.77       & 1.44 & 89.47 & 86.24 \\ 
& SFT + KTO & \textbf{91.30}    & \textbf{8.06}       & \textbf{0.65} & \textbf{91.59} & \textbf{88.23} \\ 
\bottomrule
\end{tabular}}
\caption{Results on NL to FOL conversion on the ProofWriter dataset. These models are trained on the FOLIO dataset. The ProofWriter dataset is only used for evaluation. The results are an average of 10 runs (5 in case of few-shot).}
\label{tab:results_ProofWriter}
\end{table*}

\subsection{Baselines} 
We compare our approach primarily with two baselines, i.e., the GPT3.5 and GPT4 model variants reported by LINC \citep{lincOlaussonGLZSTL23}. We refer to them as GPT3.5-LINC and GPT4-LINC respectively. We utilize the outputs for these models provided by \cite{lincOlaussonGLZSTL23}, to avoid costly experiments of GPT4-LINC. 
Finally, we evaluate off-the-shelf LLMs in their abilities to generate FOL stories by providing in-context examples.  

\paragraph{Few-shot Variants.}
\label{sec:few-shot variants}
Prior to fine-tuning, we evaluate the models' parsing abilities by using in-context examples in a few-shot setting. We use the in-context examples from FOLIO dataset \citep{lincOlaussonGLZSTL23} in \textit{1, 2, 4} and \textit{8-shot} generation tasks. Thus, the experiments on FOLIO can be considered as \textit{in-distribution} task. Meanwhile, ProofWriter and PrOntoQA are significantly different and thus requires generalizing \textit{out-of-distribution}.
\\
\textbf{Sentence-wise Translators.}~~ We also compare our models to sentence-wise parsers such as \textsc{LogicLLama} \citep{yang-etal-2024-harnessing}. We consider \textsc{GPT3.5} and \textsc{LogicLLama} in a \textit{4-shot} and \textit{5-shot} setting respectively, prompting them with NL-FOL pairs from \textsc{FOLIO}. For \textsc{LogicLLama}, the naive correction pipeline is considered due to it's improved performance. Further, we prompt \textsc{GPT3.5} with the few-shot paragraphs similar to the few-shot variants.
\\
\textbf{Fine-tuned Variants of LLMs.}~~
We use the following models for our experiments: Llama-3 \citep{grattafiori2024llama3herdmodels}, Gemma-2 \citep{gemmateam2024gemma2improvingopen}, Phi 3.5 \citep{abdin2024phi3technicalreporthighly} and Qwen-2.5 (14B).
We follow a two-stage fine-tuning approach: (i) fine-tuning these models using $\mathcal{D}_{SFT}$ dataset for SFT task and then (ii) fine-tuning them with our $\mathcal{D}_{Pref}$ dataset using preference optimization methods such as DPO \citep{rafailov2024directpreferenceoptimizationlanguage} and KTO \citep{ethayarajh2024ktomodelalignmentprospect}.

\subsection{Main Results}
\paragraph{FOLIO.} From Table \ref{tab:results_FOLIO} \& \ref{tab:results_ProofWriter}, we see that almost all of our SFT models perform close to and better than the best GPT3.5 baselines. Overall, the Phi3.5 Mini SFT version outperforms the other SFT models by atleast \textbf{4.17\%}, and GPT-3.5 LINC 8-shot by $7\%$ in the logically correct metric. The syntax error of Phi3.5 Mini SFT is also $13\%$ less than the best GPT-3.5 LINC model and $2\%$ less than the GPT-4 LINC model. 
Preference optimization with KTO also leads to further improvement in performance. We reach 61.52\% logical accuracy and 64.43\% overall F1 in FOLIO. In general, we observe that SFT + KTO always leads to improvement in performance compared to SFT, which is not the case for SFT + DPO vs. SFT.
\\
\textbf{ProofWriter and ProntoQA.}~~ We evaluate the models trained on the FOLIO dataset on the ProofWriter and ProntoQA dataset (Tabs.~\ref{tab:results_ProofWriter} \& \ref{tab:results_prontoqa}). We see a similar trend in results for these other two evaluation datasets, where SFT + KTO version of Phi3.5 Mini reaches the highest performance. We reach $65.28\%$ and $89.0\%$ logical correctness in ProofWriter and ProntoQA dataset, respectively. Llama and Gemma models show results in similar range for the ProofWriter dataset. However, for ProntoQA, the best version of Gemma significantly outperform the best LLama version.
Overall, our dual fine-tuning approach of SFT and preference optimization shows great promise towards translating NL problems into FOL. 


\section{Analysis and Ablation Studies}

\subsection{Evaluating Semantic Content of Translated FOLs}
Our automated evaluation depends on the final truth values returned by logical solvers such as \texttt{Prover9}. However, in certain cases, a logical program returning the same label as the groundtruth may not capture full semantic context. Therefore we perform both manual and automated evaluation of semantic errors.

We select samples for high performing models like \texttt{Phi3.5 Mini-KTO}, \texttt{Phi3.5 Mini SFT} and \texttt{LINC GPT3.5} where the model's output matches the ground truth evaluation of \textit{True}; and manually analyze them for semantic errors. Our analysis suggests that a high percentage of generated FOLs is the same as what we perceive in the groundtruth, with the exception that sometimes the generated predicates are overtly complex (more information) or simple (less information) compared to the groundtruth, as presented in Table ~\ref{tab:manual evaluation}. It is interesting to note that there are cases where the generated FOL is a semantic equivalent of the corresponding gold FOLs. However, we do see a gap in the model’s translation of either-or sentences, where a partly wrong translation leads to the correct label. This is a classic case of pragmatic exclusive OR  (scalar implicature) being translated as logical OR. We illustrate this as an example below. There are also cases where the model is a better representative of the NL context than the ground truth FOLs, which also suffers from this confusion.
\begin{compactitem}
\small 
    \item[] Premise: James either takes the database course or has a part-time job offered by the university.
    \item[] FOL\_gold: $((Database(James) \land \neg PartTime(James)) \lor (\neg Database(James) \land PartTime(James)))$
    \item[] FOL\_Phi: $(TakesDatabaseCourse(James) \lor HasPartTimeJob(James))$
    \item[] FOL\_GPT: $(TakeDatabaseCourse(James) \lor PartTimeJob(James))$
\end{compactitem}

\begin{table}[ht!]
\centering
\resizebox{\textwidth}{!}{
\begin{tabular}{@{}ccccccc@{}}
\toprule
Model  & \begin{tabular}[c]{@{}c@{}}Matching \\ Predicates\end{tabular} & \begin{tabular}[c]{@{}c@{}}Complicated pred.s \\ by model\end{tabular} & \begin{tabular}[c]{@{}c@{}}Simplified pred.s \\ by model\end{tabular} & \begin{tabular}[c]{@{}c@{}}Model more \\ Context-aligned\end{tabular} & \begin{tabular}[c]{@{}c@{}}Semantic \\ equivalents\end{tabular} & \begin{tabular}[c]{@{}c@{}}Either-or \\ mistakes\end{tabular} \\ \midrule
\begin{tabular}[c]{@{}c@{}}Phi3.5 Mini \\ KTO(48 samples)\end{tabular} & 17 & 9 & 9 & 4 & 1 & 6 \\ \midrule
\begin{tabular}[c]{@{}c@{}}Phi3.5 Mini \\ SFT(35 samples)\end{tabular} & 14 & 8 & 6 & - & 1 & 5 \\ \midrule
\begin{tabular}[c]{@{}c@{}}LINC GPT3.5\\ (34 samples)\end{tabular}     & 9 & 10 & 9 & - & 2 & 2  \\ \bottomrule
\end{tabular}
}
\caption{Results on manual semantic analysis of various models. Each column represents the evaluation of the FOLs generated by model as compared to the gold FOLs.}
\label{tab:manual evaluation}
\end{table}
We further use the back-translation method to judge the semantic difference between the NL story and the generated FOLs in \S\ref{sec:app:semanticeval} (Appendix), by converting the FOLs back to text. Our automated analysis shows \texttt{Llama-KTO} fares at par with \texttt{GPT3.5-LINC} in this regard, showing high levels of semantic preservation in the generated FOLs.

\subsection{Qualitative analysis}
From the results reported in Tabs.~ \ref{tab:results_FOLIO}, \ref{tab:results_ProofWriter} \& \ref{tab:results_prontoqa} (Appendix), it is evident that our fine-tuned models are significantly better in all evaluation criteria. We further explore \textit{How the generations from fine-tuned models differ from other neurosymbolic models?} We focus on comparing Llama-3 8b instruct+SFT+KTO model vs. Phi-3.5 Mini+SFT+KTO as their performance exceeds other models. We also compare our models to LINC GPT3.5 to qualitatively examine any improvements. 
Qualitatively, we find that both the models share few similarities and dissimilarities in their generations and highlight some key findings from both. We also compare the model errors with the different failure modes for LINC reported in \cite{lincOlaussonGLZSTL23}. We use the categories L1, L2 and L3 to denote implicit information (not mentioned in text but required) loss, explicit information (required and mentioned in text) mistakes, and syntax errors correspondingly. 
\\
\textbf{Similarity $S_1$: Lack of consistent usage of predicates.}~~
Often, the models use different predicate to convey the same meaning in subsequent sentences leading to loss of information to Prover9. Nevertheless, finetuned Llama-3 suffers less from this type of errors since we see more consistent usage of predicates in Llama. For example\footnote{Please note that LINC GPT3.5 uses $Flies$ predicate instead of $FliesTo$ and $FliesFrom$, we omitted them in the example to avoid complications}, in the snippet below: 
\begin{compactitem}
\small
    \item[] Premise 1: Susan flies to LGA airport.
    \item[] FOL: $FliesTo(Susan, LGAAirport)$
    \item[] \textcolor{red}{Premise 2: The departure and arrival can not be the same airport. }
    \item[] \textcolor{red}{FOL\_Llama: $\neg EqualAirports(Daniel, Susan)$}
    \item[] \textcolor{red}{FOL\_Phi: $\neg(DepartFrom(x) \land ArriveAt(x))$}
    \item[] \textcolor{red}{FOL\_LINC: $\forall x \forall y (Departure(x) \land Arrival(y) \land \neg SameAirport(x, y))$}
    \item[] Premise 3: John flies from LGA airport.
    \item[] FOL: $FliesFrom(John, LGAAirport)$
    \item[] Premise 4: Susan flies from LGA airport.
    \item[] FOL: $FliesFrom(Susan, LGAAirport)$
\end{compactitem}

\textbf{Similarity $S_2$: Both models suffer with complex logic.}
Both models fare poorly with large sentences with more complex logic. However, \texttt{LINC GPT3.5} excels in this category while not suffering from such flaws. In the example below, both the models suffer in the same way confusing neither-nor logic while LINC produces the gold FOL: 
\begin{compactitem}
    \item[] TEXT: If Rock is neither a fly nor a bird, then Rock neither flies nor breathes.
    \item[] \textcolor{red}{FOL: $\neg((Fly(Rock) \lor Bird(Rock)) \Rightarrow (\neg Fly(Rock) \land -Breathes(Rock)))$}
    \item[] \textcolor{teal}{FOL\_LINC: $(\neg Fly(Rock) \& \neg Bird(Rock)) \Rightarrow (\neg Fly(Rock) \& \neg Breathe(Rock))$}
\end{compactitem}

\textbf{Similarity $S_3$: Problem with Uncertain labels.}~~
We find the accuracies related to \textit{Uncertain} label highly unreliable due to the high number of ways in which \textit{Uncertain} label can be reached. We have found the following methods the LLM models utilize in order to unexpectedly reach the \textit{Uncertain} label: 1) The LLM has an inconsistent usage of predicates, like $S_1$. 2) Incorrect translation of a single FOL in the story causing loss of information. 

\paragraph{Error modes of LINC.} According to \cite{lincOlaussonGLZSTL23}, there are mainly 3 modes of failure for neurosymbolic solvers like \textsc{LINC}. Using FOLIO (validation) problems, we manually annotate the (incorrect) responses from our best models (\textsc{Llama-3} and \textsc{Phi3} with SFT+KTO) with the three labels (L1, L2, L3) to investigate whether preference optimization helps alleviate these issues. We further compare with contemporary neurosymbolic provers, mainly Logic-LM \citep{logiclmemnlpPanAWW23} and LINC \citep{lincOlaussonGLZSTL23}.
Table ~\ref{tab:FOLIO Errors} shows our results. We see a huge decrease in Syntax errors (L3) in our models compared to these baselines. With an 83.33\% decrease from Logic-LM and 61.11\% decrease from LINC, we also see a huge jump in exectutability of the programs generated by our models. We see a decrease in implicit information loss across various failures. Implicit information refers to commonsense knowledge (or ontological knowledge) required to derive the conclusion. We see that \textsc{LogicPO}-finetuned models better consolidate such information, sometimes adding extraneous information. However, we fallback on explicit information (L2) leading to more failures than LINC. This is possibly because our models are using larger predicates that contain more information embedded in FOLs. Overall, our models show improvement over other neurosymbolic approaches with less syntactic and logical errors, yet there are failures that can be overcome. 

\begin{table}[t!]
\small
\centering
\resizebox{0.5\textwidth}{!}{
\begin{tabular}{l|cccc}
\toprule[1pt]
Model & L1 & L2 & L3 & \begin{tabular}[c]{@{}c@{}}Wrong\\ Translation\end{tabular} \\ \hline 
Logic-LM GPT3.5 & 18 & 20 & 84 & 4 \\ 
LINC GPT3.5 & 22 & 13 & 36 & 1 \\
Llama-3 8B Instruct & 10 & 15 & 14 & 9 \\
Phi-3.5 Mini Instruct & 13 & 20 & 14 & 1 \\ \bottomrule
\end{tabular}
}
\caption{\small{Number of instances in FOLIO, with errors corresponding to errors from LINC. L1: Implicit Information Loss, L2: Explicit information errors, L3: Syntax Errors}}
\label{tab:FOLIO Errors}
\end{table}

\paragraph{Analysis of NL to FOL Conversion across Input Lengths.}

We analyze how well the fine-tuned models convert NL to FOL stories across different input context sizes in Table~\ref{tab:length_analysis} (Appendix). We group the FOLIO dataset into three categories -- instances with small (1-2 sentences), medium (3-5 sentences) and large context (more than 5 sentences). Llama and Gemma models shows monotonically decreasing performance as we increase the context length. Interestingly, the Phi models do almost similarly in the small and medium context instances, which is not the case for the other two models.

\section{Conclusion}
In this work, we present an efficient method for improving logical reasoning of LLMs through preference optimization on a synthetically generated dataset. We propose a way to synthesize a large ($26k$) dataset, which contains correct and incorrect FOL programs for each natural language reasoning problem (context and question).  Our experiments show that preference optimization on this dataset leads to significant performance gains in all of our evaluation criteria. Furthermore, carrying out a qualitative and quantitative analysis of our models shows the various advantages and shortcomings of our approach. This work thus shows promise in the field of LLMs as parsers through preference optimization. Paving the way for future work on continual fine-tuning of neurosymbolic solvers for logical reasoning. 

\section*{Limitations}
Our work is among the first ones which attempts to convert natural language reasoning problems holistically to an equivalent logical representation in First Order Logic. The primary limitations of the work is as follows. 
\\\noindent
1) At various stages of dataset creation, we depend on the predicted logical label. However, it is not guaranteed that if the logical label is correct, the program will also be correct. While we attempt to evaluate the semantic content, this is clearly an open problem and requires further exploration. This is a problem that plagues other similar efforts (such as ProverQA).
\\\noindent
2) We only explore English-FOL as a representative natural-formal language pair combinations. Provided the current failure modes of LLMs, it is probable that parsing errors will be higher as we change to even low-resource formal languages or low-resource natural language. Many low-resource formal languages have been shown to be useful such as LEAN for mathematical theorem proving (was used for GPT-4's math olympiad work). One can possibly adopt our framework for generalizing to such languages as well.

\bibliography{colm2025_conference,anthology,latex/custom}
\bibliographystyle{colm2025_conference}

\appendix
\section{Appendix}
\subsection{Few-shot Baseline prompts}
For our baseline models, we evaluate off-the-shelf models in a few-shot setting. As explained in section~\ref{sec:few-shot variants}, we use \textsc{FOLIO} excerpts in our $2$ and $4$ -shot generation tasks. Note that we use the same examples for evaluation across various datasets. For each of our examples, we provide information about premises, conclusion, premises-FOL and conclusion-FOL which are a part of the \textsc{FOLIO} dataset. Inspired from \cite{lincOlaussonGLZSTL23}, we use HTML-style tags to wrap information about each part of the example. First, we provide information about the premise story wrapped in \texttt{<PREMISES>...</PREMISES>} followed by conclusion sentence in \texttt{<CONCLUSION>...</CONCLUSION>} tags. We then use the \texttt{<EVALUATE>} tag to indicate the LLMs to start generation, and \texttt{</EVALUATE>} indicating the end of generation. We alternate between the text and it's corresponding FOL translation generated indicated by \texttt{TEXT:} and \texttt{FOL:} respectively. The conclusion sentence and it's translation are appended to the end of the story in the evaluate section itself to eliminate any potential confusion. \\
\paragraph{2-shot prompt}
\begin{longtable}{@{}|p\textwidth|@{}}

\toprule
\textcolor{blue}{You are an expert in working with first-order logic (FOL) problems.} \\
\textcolor{blue}{You will be given a context with a set of premise sentences and a single conclusion sentence. }\\
\textcolor{blue}{Your task is to translate each of the premise sentences and the conclusion sentence into FOL expressions,} \\
\textcolor{blue}{so that the expressions can be evaluated by a theorem solver to determine whether the conclusion follows from the premise sentences.}\\ 
\textcolor{blue}{Expressions should be adhere to the format of the Python NLTK package logic module.}\\ 
\textcolor{blue}{Here are some examples of the task:}\
\ 
\\ \textcolor{red}{Example 1:}\\ 
\textcolor{teal}{\textless{}PREMISES\textgreater}\\ 
\textcolor{teal}{All dispensable things are environment-friendly.}\\ 
\textcolor{teal}{All woodware is dispensable.}\\ 
\textcolor{teal}{All paper is woodware.}\\ 
\textcolor{teal}{No good things are bad.}\\ 
\textcolor{teal}{All environment-friendly things are good.}\\ 
\textcolor{teal}{A worksheet is either paper or is environment-friendly.}\\ 
\textcolor{teal}{\textless{}/PREMISES\textgreater}\\ 

\textcolor{violet}{\textless{}CONCLUSION\textgreater}\\ 
\textcolor{violet}{A worksheet is not dispensable.}\\ 
\textcolor{violet}{\textless{}/CONCLUSION\textgreater}\\ 

\textcolor{purple}{\textless{}EVALUATE\textgreater{}}\\
\textcolor{teal}{TEXT:	All dispensable things are environment-friendly.}\\ 
\textcolor{olive}{FOL:	all x. (Dispensable(x) -\textgreater EnvironmentFriendly(x))}\\ 
\textcolor{teal}{TEXT:	All woodware is dispensable.}\\ 
\textcolor{olive}{FOL:	all x. (Woodware(x) -\textgreater Dispensable(x))}\\ 
\textcolor{teal}{TEXT:	All paper is woodware.}\\ 
\textcolor{olive}{FOL:	all x. (Paper(x) -\textgreater Woodware(x))}\\ 
\textcolor{teal}{TEXT:	No good things are bad.}\\ 
\textcolor{olive}{FOL:	all x. (Good(x) -\textgreater -Bad(x))}\\ 
\textcolor{teal}{TEXT:	All environment-friendly things are good.}\\ 
\textcolor{olive}{FOL:	all x. (EnvironmentFriendly(x) -\textgreater Good(x))}\\ 
\textcolor{teal}{TEXT:	A worksheet is either paper or is environment-friendly.}\\ 
\textcolor{olive}{FOL:	((Paper(Worksheet) \& -EnvironmentFriendly(Worksheet)) | (-Paper(Worksheet) \& EnvironmentFriendly(Worksheet)))}\\ 
\textcolor{violet}{TEXT:	A worksheet is not dispensable.}\\ 
\textcolor{violet}{FOL:	-Dispensable(Worksheet)}\\ 
\textcolor{purple}{\textless{}/EVALUATE\textgreater}\\ \\ 

\textcolor{red}{Example 2:}\\ 
\textcolor{teal}{\textless{}PREMISES\textgreater} \\ 
\textcolor{teal}{A La Liga soccer team ranks higher than another if it receives more points.}\\ 
\textcolor{teal}{If two La Liga soccer teams recieve the same points, }\\
\textcolor{teal}{the team which recieves more points from the games between the two teams ranks higher.}\\ 
\textcolor{teal}{Real Madrid and Barcelona are both La Liga soccer teams.}\\ 
\textcolor{teal}{In La Liga 2021-2022, Real Madrid recieves 86 points and Barcelon recieves 73 points.}
\\ 
\textcolor{teal}{In La Liga 2021-2022, Real Madrid and Barcelona both recieve 3 points from the games between them.}\\ 
\textcolor{teal}{\textless{}/PREMISES\textgreater} \\ 

\textcolor{violet}{\textless{}CONCLUSION\textgreater} \\ 
\textcolor{violet}{In La Liga 2021-2022, Real Madrid ranks higher than Barcelona.}\\ 
\textcolor{violet}{\textless{}/CONCLUSION\textgreater} \\ 

\textcolor{purple}{\textless{}EVALUATE\textgreater{}} \\
\textcolor{teal}{TEXT:	A La Liga soccer team ranks higher than another if it receives more points.}\\ 
\textcolor{olive}{FOL:	all x. all y. (LaLiga(x) \& LaLiga(y) \& MorePoints(x, y) -\textgreater HigherRank(x, y))}\\ 
\textcolor{teal}{TEXT:	If two La Liga soccer teams recieve the same points, }\\
\textcolor{teal}{the team which recieves more points from the games between the two teams ranks higher.}\\ 
\textcolor{olive}{FOL:	all x. all y. (LaLiga(x) \& LaLiga(y) \& -MorePoints(x, y) \& -MorePoints(y, x) }\\ 
\textcolor{olive}{\& MorePointsInGameBetween(x, y) -\textgreater HigherRank(x, y))}\\ 
\textcolor{teal}{TEXT:	Real Madrid and Barcelona are both La Liga soccer teams.}\\ 
\textcolor{olive}{FOL:	LaLiga(RealMadrid) \& LaLiga(Barcelona)}\\ 
\textcolor{teal}{TEXT:	In La Liga 2021-2022, Real Madrid recieves 86 points and Barcelon recieves 73 points.}\\ 
\textcolor{olive}{FOL:	MorePoints(RealMadrid, Barcelona)}\\ 
\textcolor{teal}{TEXT:	In La Liga 2021-2022, Real Madrid and Barcelona both recieve 3 points from the games between them.}\\ 
\textcolor{olive}{FOL:	-MorePointsInGameBetween(RealMadrid, Barcelona) \& -MorePointsInGameBetween(Barcelona, RealMadrid)}\\ 
\textcolor{violet}{TEXT:	In La Liga 2021-2022, Real Madrid ranks higher than Barcelona.}\\ 
\textcolor{violet}{FOL:	HigherRank(RealMadrid, Barcelona)}\\ 
\textcolor{purple}{\textless{}/EVALUATE\textgreater} \\ \\ \\

\textcolor{blue}{Notice the output inside the \textless{}EVALUATE\textgreater and s\textless{}/EVALUATE\textgreater block. We have taken each sentence from our premise and conclusion and }\\
\textcolor{blue}{converted it to the corresponding FOL expression. The lines starting with TEXT: copies the original sentence from our context. The lines starting}\\ \textcolor{blue}{with FOL: shows the corresponding FOL form.}\\ \\ 

\textcolor{blue}{Can you now generate the FOL expressions for the following example, maintaining the format shown earlier. Do not generate any explanations.}\\ 
\textcolor{teal}{\textless{}PREMISES\textgreater} \\ 
\textcolor{teal}{...} \\
\textcolor{teal}{\textless{}\textbackslash{}PREMISES\textgreater}\\ \textcolor{purple}{\textless{}EVALUATE\textgreater{}}\\
\textcolor{purple}{...} \\
\textcolor{purple}{\textless{}\textbackslash{}EVALUATE\textgreater{}}\\
\bottomrule
\caption{\label{tabl: 2-prompt}2-shot prompt for baseline experiments}
\end{longtable}

\paragraph{4-shot-prompt}
\begin{longtable}{@{}|p\textwidth|@{}}
\toprule
\textcolor{blue}{You are an expert in working with first-order logic (FOL) problems.} \\
\textcolor{blue}{You will be given a context with a set of premise sentences and a single conclusion sentence. }\\
\textcolor{blue}{Your task is to translate each of the premise sentences and the conclusion sentence into FOL expressions,} \\
\textcolor{blue}{so that the expressions can be evaluated by a theorem solver to determine whether the conclusion follows from the premise sentences.}\\ 
\textcolor{blue}{Expressions should be adhere to the format of the Python NLTK package logic module.}\\ 
\textcolor{blue}{Here are some examples of the task:}\
\ 
\\ \textcolor{red}{Example 1:}\\ 
\textcolor{teal}{\textless{}PREMISES\textgreater}\\ 
\textcolor{teal}{All dispensable things are environment-friendly.}\\ 
\textcolor{teal}{All woodware is dispensable.}\\ 
\textcolor{teal}{All paper is woodware.}\\ 
\textcolor{teal}{No good things are bad.}\\ 
\textcolor{teal}{All environment-friendly things are good.}\\ 
\textcolor{teal}{A worksheet is either paper or is environment-friendly.}\\ 
\textcolor{teal}{\textless{}/PREMISES\textgreater}\\ 

\textcolor{violet}{\textless{}CONCLUSION\textgreater}\\ 
\textcolor{violet}{A worksheet is not dispensable.}\\ 
\textcolor{violet}{\textless{}/CONCLUSION\textgreater}\\ 

\textcolor{purple}{\textless{}EVALUATE\textgreater{}}\\
\textcolor{teal}{TEXT:	All dispensable things are environment-friendly.}\\ 
\textcolor{olive}{FOL:	all x. (Dispensable(x) -\textgreater EnvironmentFriendly(x))}\\ 
\textcolor{teal}{TEXT:	All woodware is dispensable.}\\ 
\textcolor{olive}{FOL:	all x. (Woodware(x) -\textgreater Dispensable(x))}\\ 
\textcolor{teal}{TEXT:	All paper is woodware.}\\ 
\textcolor{olive}{FOL:	all x. (Paper(x) -\textgreater Woodware(x))}\\ 
\textcolor{teal}{TEXT:	No good things are bad.}\\ 
\textcolor{olive}{FOL:	all x. (Good(x) -\textgreater -Bad(x))}\\ 
\textcolor{teal}{TEXT:	All environment-friendly things are good.}\\ 
\textcolor{olive}{FOL:	all x. (EnvironmentFriendly(x) -\textgreater Good(x))}\\ 
\textcolor{teal}{TEXT:	A worksheet is either paper or is environment-friendly.}\\ 
\textcolor{olive}{FOL:	((Paper(Worksheet) \& -EnvironmentFriendly(Worksheet)) | (-Paper(Worksheet) \& EnvironmentFriendly(Worksheet)))}\\ 
\textcolor{violet}{TEXT:	A worksheet is not dispensable.}\\ 
\textcolor{violet}{FOL:	-Dispensable(Worksheet)}\\ 
\textcolor{purple}{\textless{}/EVALUATE\textgreater}\\ \\ 

\textcolor{red}{Example 2:}\\ 
\textcolor{teal}{\textless{}PREMISES\textgreater} \\ 
\textcolor{teal}{A La Liga soccer team ranks higher than another if it receives more points.}\\ 
\textcolor{teal}{If two La Liga soccer teams recieve the same points, }\\
\textcolor{teal}{the team which recieves more points from the games between the two teams ranks higher.}\\ 
\textcolor{teal}{Real Madrid and Barcelona are both La Liga soccer teams.}\\ 
\textcolor{teal}{In La Liga 2021-2022, Real Madrid recieves 86 points and Barcelon recieves 73 points.}
\\ 
\textcolor{teal}{In La Liga 2021-2022, Real Madrid and Barcelona both recieve 3 points from the games between them.}\\ 
\textcolor{teal}{\textless{}/PREMISES\textgreater} \\ 

\textcolor{violet}{\textless{}CONCLUSION\textgreater} \\ 
\textcolor{violet}{In La Liga 2021-2022, Real Madrid ranks higher than Barcelona.}\\ 
\textcolor{violet}{\textless{}/CONCLUSION\textgreater} \\ 

\textcolor{purple}{\textless{}EVALUATE\textgreater{}} \\
\textcolor{teal}{TEXT:	A La Liga soccer team ranks higher than another if it receives more points.}\\ 
\textcolor{olive}{FOL:	all x. all y. (LaLiga(x) \& LaLiga(y) \& MorePoints(x, y) -\textgreater HigherRank(x, y))}\\ 
\textcolor{teal}{TEXT:	If two La Liga soccer teams recieve the same points, }\\
\textcolor{teal}{the team which recieves more points from the games between the two teams ranks higher.}\\ 
\textcolor{olive}{FOL:	all x. all y. (LaLiga(x) \& LaLiga(y) \& -MorePoints(x, y) \& -MorePoints(y, x) }\\ 
\textcolor{olive}{\& MorePointsInGameBetween(x, y) -\textgreater HigherRank(x, y))}\\ 
\textcolor{teal}{TEXT:	Real Madrid and Barcelona are both La Liga soccer teams.}\\ 
\textcolor{olive}{FOL:	LaLiga(RealMadrid) \& LaLiga(Barcelona)}\\ 
\textcolor{teal}{TEXT:	In La Liga 2021-2022, Real Madrid recieves 86 points and Barcelon recieves 73 points.}\\ 
\textcolor{olive}{FOL:	MorePoints(RealMadrid, Barcelona)}\\ 
\textcolor{teal}{TEXT:	In La Liga 2021-2022, Real Madrid and Barcelona both recieve 3 points from the games between them.}\\ 
\textcolor{olive}{FOL:	-MorePointsInGameBetween(RealMadrid, Barcelona) \& -MorePointsInGameBetween(Barcelona, RealMadrid)}\\ 
\textcolor{violet}{TEXT:	In La Liga 2021-2022, Real Madrid ranks higher than Barcelona.}\\ 
\textcolor{violet}{FOL:	HigherRank(RealMadrid, Barcelona)}\\ 
\textcolor{purple}{\textless{}/EVALUATE\textgreater} \\  \\ 

\textcolor{red}{Example 3:}\\ 
\textcolor{teal}{\textless{}PREMISES\textgreater} \\ 
\textcolor{teal}{All athletes are good at sports.} \\
\textcolor{teal}{All Olympic gold medal winners are good athletes.} \\
\textcolor{teal}{No scientists are good at sports.} \\
\textcolor{teal}{All Nobel laureates are scientists.} \\
\textcolor{teal}{Amy is good at sports or Amy is an Olympic gold medal winner.} \\
\textcolor{teal}{If Amy is not a Nobel laureate, then Amy is not an Olympic gold medal winner.} \\
\textcolor{teal}{\textless{}/PREMISES\textgreater} \\ 

\textcolor{violet}{\textless{}CONCLUSION\textgreater}\\ 
\textcolor{violet}{If Amy is not an Olympic gold medal winner, then Amy is a Nobel laureate.}
\textcolor{violet}{\textless{}/CONCLUSION\textgreater}\\ 
\textcolor{purple}{\textless{}EVALUATE\textgreater{}} \\
\textcolor{teal}{TEXT:	All athletes are good at sports.}\\
\textcolor{olive}{FOL: all x. (Athlete(x) -\textgreater GoodAtSports(x))} \\
\textcolor{teal}{TEXT:	All Olympic gold medal winners are good athletes.}\\
\textcolor{olive}{FOL: all x. (OlympicGoldMedalWinner(x) -\textgreater Athlete(x)) } \\
\textcolor{teal}{TEXT:	No scientists are good at sports.}\\
\textcolor{olive}{FOL:	all x. (Scientist(x) -\textgreater -GoodAtSports(x))} \\
\textcolor{teal}{TEXT:	All Nobel laureates are scientists.}\\
\textcolor{olive}{FOL:	all x. (NobelLaureate(x) -\textgreater Scientist(x))} \\
\textcolor{teal}{TEXT:	Amy is good at sports or Amy is an Olympic gold medal winner.}\\
\textcolor{olive}{FOL:	GoodAtSports(Amy) | OlympicGoldMedalWinner(Amy)} \\
\textcolor{teal}{TEXT:	If Amy is not a Nobel laureate, then Amy is not an Olympic gold medal winner.}\\
\textcolor{olive}{FOL:	-NobelLaureate(Amy) -\textgreater -OlympicGoldMedalWinner(Amy)} \\
\textcolor{violet}{TEXT:	If Amy is not an Olympic gold medal winner, then Amy is a Nobel laureate.}\\
\textcolor{violet}{FOL:	-OlympicGoldMedalWinner(Amy) -\textgreater NobelLaureate(Amy)} \\ 
\textcolor{purple}{\textless{}/EVALUATE\textgreater{}} \\ \\

\textcolor{red}{Example 4:}\\ 
\textcolor{teal}{\textless{}PREMISES\textgreater} \\ 
\textcolor{teal}{All people who are respected by others are people who contribute to the country.}\\
\textcolor{teal}{If a person is respected by others, then he/she contributes to the country.}\\
\textcolor{teal}{All people available to have a visit without any fees are those respected by others.}\\
\textcolor{teal}{All Customers who once served in the army are available to have a visit without any fees.}\\
\textcolor{teal}{All people who once were sentenced for thief stayed in prison for some time.}\\
\textcolor{teal}{All people who once stayed in prison for some time have a bad record in the local state.}\\
\textcolor{teal}{James was either once sentenced for thief or stayed in prison for some time.}\\
\textcolor{teal}{James is either with a bad record in the local state or respected by others.}\\
\textcolor{teal}{\textless{}/PREMISES\textgreater} \\ 

\textcolor{violet}{\textless{}CONCLUSION\textgreater}\\ 
\textcolor{violet}{James contributes to the country.}\\
\textcolor{violet}{\textless{}/CONCLUSION\textgreater}\\ 

\textcolor{purple}{\textless{}EVALUATE\textgreater{}} \\ \textcolor{teal}{TEXT:	All people who are respected by others are people who contribute to the country.}\\
\textcolor{olive}{FOL:	all x. (Respected(x) -\textgreater ContributeToCountry(x))}\\
\textcolor{teal}{TEXT:	If a person is respected by others, then he/she contributes to the country.}\\
\textcolor{olive}{FOL:	all x. (Respected(x) -\textgreater ContributeToCountry(x))}\\
\textcolor{teal}{TEXT:	All people available to have a visit without any fees are those respected by others.}\\
\textcolor{olive}{FOL:	all x. (HaveVisitWithoutAnyFees(x) -\textgreater Respected(x))}\\
\textcolor{teal}{TEXT:	All Customers who once served in the army are available to have a visit without any fees.}\\
\textcolor{olive}{FOL:	all x. (Army(x) -\textgreater HaveVisitWithoutAnyFees(x))}\\
\textcolor{teal}{TEXT:	All people who once were sentenced for thief stayed in prison for some time.}\\
\textcolor{olive}{FOL:	all x. (Prison(x) -\textgreater BadRecord(x))}\\
\textcolor{teal}{TEXT:	All people who once stayed in prison for some time have a bad record in the local state.}\\
\textcolor{olive}{FOL:	all x. (Prison(x) -> BadRecord(x))}\\
\textcolor{teal}{TEXT:	James was either once sentenced for thief or stayed in prison for some time.}\\
\textcolor{olive}{FOL:	((Thief(James) \& -Prison(James)) | (-Thief(James) \& Prison(James)))}\\
\textcolor{teal}{TEXT:	James is either with a bad record in the local state or respected by others.}\\
\textcolor{olive}{FOL:	((BadRecord(James) \& -Respected(James)) | (-BadRecord(James) \& Respected(James)))}\\
\textcolor{violet}{TEXT:  James contributes to the country.}\\
\textcolor{violet}{FOL:	ContributeToCountry(James)}\\
\textcolor{purple}{\textless{}/EVALUATE\textgreater{}} \\ \\

\textcolor{blue}{Notice the output inside the \textless{}EVALUATE\textgreater and \textless{}/EVALUATE\textgreater block. We have taken each sentence from our premise and conclusion and }\\
\textcolor{blue}{converted it to the corresponding FOL expression. The lines starting with TEXT: copies the original sentence from our context. The lines starting}\\ \textcolor{blue}{with FOL: shows the corresponding FOL form.}\\ \\ 

\textcolor{blue}{Can you now generate the FOL expressions for the following example, maintaining the format shown earlier. Do not generate any explanations.}\\ 
\textcolor{teal}{\textless{}PREMISES\textgreater} \\ 
\textcolor{teal}{...} \\
\textcolor{teal}{\textless{}\textbackslash{}PREMISES\textgreater}\\ \textcolor{purple}{\textless{}EVALUATE\textgreater{}}\\
\textcolor{purple}{...} \\
\textcolor{purple}{\textless{}\textbackslash{}EVALUATE\textgreater{}}\\
\bottomrule
\caption{4-shot prompt for baseline experiments}
\label{tabl: 4-prompt}
\end{longtable}

\section{Additional Results}
We provide additional results of NL reasoning problem to FOL translation on PrOntoQA validation dataset in Table~\ref{tab:results_prontoqa}.

\begin{table*}[!ht]
\small
\centering
\resizebox{0.9\textwidth}{!}{
\begin{tabular}{@{}ll|ccc|cc@{}}
\toprule
\multirow{2}{*}{\textbf{Model}} & \multirow{2}{*}{\textbf{Setting}} & \multicolumn{2}{c}{\textbf{Logically}} & \textbf{Syntax} & \multicolumn{2}{c}{\textbf{F1}} \\ 
& & \textbf{Correct (↑)} & \textbf{Incorrect} & \textbf{Error (↓)} & \textbf{Overall (↑)} & \textbf{True Label (↑)}\\
\midrule

\multirow{3}{*}{Llama-3 8B Instruct} 
     & 2-shot & 83.40 & 5.4 & 11.2 & 89.87 & 89.94 \\
     & 4-shot & 45.00 & 42.8 & 12.2 & 33.31 & 64.56 \\
     & SFT & 27.2 & 65.0 & 7.8 & 41.79 & 45.98 \\
     & SFT + DPO & 45.2 & 48.4 & 6.4 & 61.48 & 62.50 \\
     & SFT + KTO & 47.4 & 45.4 & 7.2 & 63.18 & 64.14 \\
\midrule
\multirow{3}{*}{Gemma-2 2B Instruct} 
     & 2-shot & 21.12 & 49 & 29.88 & 27.52 & 46.53 \\
     & 4-shot & 36.25 & 37.5 & 26.25 & 28.85 & 59.18 \\
     & SFT & 49.2 & 44.6 & 6.2 & 64.14 & 61.77 \\
     & SFT + DPO & 40.4 & 29.4 & 30.2 & 56.19 & 55.53 \\
     & SFT + KTO & 56.4 & 33.4 & 10.2 & 70.23 & 68.45 \\
\midrule
\multirow{3}{*}{Phi-3.5 Mini Instruct (4B)} 
     & 2-shot & 66.00 & 27.77 & 15.4 & 76.80 & 78.32 \\
     & 4-shot & 52.60 & 18.40 & 29.00 & 68.23 & 72.15 \\
     & SFT & 87.8 & 11.6 & \textbf{0.6} & 92.81 & 92.65 \\
     & SFT + DPO & 80.2 & 18.6 & 1.2 & 88.32 & 87.37 \\
     & SFT + KTO & \textbf{89.0} & 10.4 & \textbf{0.6} & \textbf{93.68} & \textbf{93.93} \\
\midrule
\multirow{3}{*}{Qwen-2.5 Instruct (14B)} 
     & 2-shot & 97.40 & 2.6 & 0.0 & 98.68 & 98.62 \\
     & SFT       & 81.13                      & 18.17                        & 0.70                  & 89.00               & 89.57                  \\ 
& SFT + DPO & \textbf{92.57}      
& \textbf{7.27}                & \textbf{0.17}         & 95.04               & 94.96                  \\ 
& SFT + KTO & 91.17                      & 8.37                         & 0.47                  & \textbf{95.34}      & \textbf{96.21}          \\
\bottomrule
\end{tabular}}
\caption{Results on NL to FOL conversion on the ProntoQA validation dataset. These models are trained on the FOLIO dataset. The ProntoQA dataset is only used for evaluation. The results are an average of 10 runs (5 in case of few-shot).}
\label{tab:results_prontoqa}
\end{table*}

\subsection{Additional ProverQA Results} 
\label{sec:app:addl:proverqa}
The ProverQA dataset \citep{qi2025largeproverqa} is a high-quality and diverse FOL reasoning benchmark generated using the Llama3.1-70B-Instruct model. The dataset features complex rules and faithful intermediate reasoning steps. The dataset has three splits depending on the length of the reasoning chain: easy (1-2 steps), medium (3-5 steps), and hard (6-9 steps). Each split contains 500 instances. 
They show that finetuning Llama3.1-8B-Instruct on this training dataset produces superior accuracy boosts on both in-distribution and out-of-distribution test sets -- obtaining an accuracy of $88.40$, $65.67$ and $68.57$ for ProntoQA, ProofWriter and FOLIO respectively. Our best finetuned models obtain similar scores (\textsc{F1}) as shown across various tables (Tabs.~\ref{tab:results_prontoqa}, \ref{tab:results_ProofWriter} \& \ref{tab:results_FOLIO}), bolstering the generalizability of \textsc{LogicPO}. It is also well known that \textsc{F1} scores provide a more comprehensive view of model performance.  Furthermore, we provide inference results on different splits of ProverQA, with all the trained models on \textsc{LogicPO} in Table~\ref{tab:results_proverqa} \& ~\ref{tab:proverqa:qwen}. We achieve the highest syntactic correctness again with the Phi3.5 models. As the process of ProverQA creation was quite different, the high improvement of the preference optimized models underlines the generalizability of our \textsc{LogicPO} dataset.

\begin{table*}[!ht]
\small
\centering
\resizebox{0.9\textwidth}{!}{
\begin{tabular}{@{}ll|ccc|ccc|ccc}
\toprule
\multirow{2}{*}{\textbf{Model}} & \multirow{2}{*}{\textbf{Setting}} & \multicolumn{3}{c}{\textbf{Easy Set}} & \multicolumn{3}{c}{\textbf{Medium Set}} & \multicolumn{3}{c}{\textbf{Hard Set}} \\ 
& & \textbf{Correct (↑)} & \textbf{Incorrect} & \textbf{Error (↓)} & \textbf{Correct (↑)} & \textbf{Incorrect} & \textbf{Error (↓)} & \textbf{Correct (↑)} & \textbf{Incorrect} & \textbf{Error (↓)}\\
\midrule

\multirow{3}{*}{Llama-3 8B Instruct} 
     & SFT & 57.2 & 36.8 & 6.0 & 34.1 & 46.9 & 19.0 & 24.1 & 47.8 & 28.1\\
     & SFT + DPO & 48.2 & 32.3 & 19.5 & 20.6 & 23.1 & 56.3 & 7.4 & 15.1 & 77.5\\
     & SFT + KTO & 55.9 & 32.8 & 11.2 & 30.7 & 34.6 & 34.7 & 14.5 & 30.6 & 54.9 \\
\midrule
\multirow{3}{*}{Gemma-2 2B Instruct} 
     & SFT & 53.5 & 33.7 & 12.8 & 35.9 & 42.0 & 22.1 & 23.7 & 46.8 & 29.5 \\
     & SFT + DPO & 41.4 & 16.2 & 42.4 & 8.6 & 5.6 & 85.8 & 1.0 & 1.9 & 97.1 \\
     & SFT + KTO & 54.4 & 25.9 & 19.7 & 28.0 & 27.0 & 45.0 & 14.3 & 22.5 & 63.2 \\
\midrule
\multirow{3}{*}{Phi-3.5 Mini Instruct (4B)} 
     & SFT & 75.1 & 21.7 & \textbf{3.2} & 54.3 & 38.4 & \textbf{7.3} & 32.4 & 44.7 & 22.9 \\
     & SFT + DPO & 76.2 & 16.9 & 6.9 & 52.1 & 38.4 & 9.5 & \textbf{33.1} & 45.7 & \textbf{21.2} \\
     & SFT + KTO & \textbf{80.4} & \textbf{15.4} & 4.2 & \textbf{57.9} & 33.9 & 8.2 & 28.5 & 44.1 & 27.4 \\
\bottomrule
\end{tabular}}
\caption{Results on NL to FOL conversion on the easy, medium and hard splits on the ProverQA dataset. These models are trained on the FOLIO dataset. The ProverQA dataset is only used for evaluation. The results are an average of 5 runs.}
\label{tab:results_proverqa}
\end{table*}

\begin{table}[!htpb]
\small
\centering
\resizebox{0.9\textwidth}{!}{%
\begin{tabular}{@{}cccccc@{}}
\toprule
\textbf{} & \textbf{Correct} ($\uparrow$) & \textbf{Incorrect} & \textbf{Syntax Error} ($\downarrow$) & \textbf{Overall F1} ($\uparrow$) & \textbf{True F1}($\uparrow$) \\ \midrule
          & \multicolumn{5}{c}{ProverQA (Easy)}                                                                                              \\ \midrule
SFT       & 82.40                      & 16.53                        & 1.07                  & 82.96               & 84.92                  \\
SFT + DPO & \textbf{91.93}             & \textbf{7.23}                & 0.83         & 92.37      & 93.25         \\
SFT + KTO & 90.70                      & 8.77                         & \textbf{0.53}         & \textbf{91.04}      & \textbf{92.24}         \\ \midrule
          & \multicolumn{5}{c}{ProverQA (Medium)}                                                                                            \\ \midrule
SFT       & 71.83                      & 24.40                        & 3.77                  & 72.72               & 69.30                  \\
SFT + DPO & 79.47                      & 17.83                        & 2.70                  & 80.69               & 79.47                  \\
SFT + KTO & \textbf{80.57}             & \textbf{17.83}               & \textbf{1.60}         & \textbf{81.25}      & \textbf{82.01}         \\ \midrule
          & \multicolumn{5}{c}{ProverQA (Hard)}                                                                                              \\ \midrule
SFT       & 42.77                      & 34.27                        & 22.97                 & 45.34               & 39.84                  \\
SFT + DPO & 45.43                      & \textbf{30.53}                        & 24.03                 & 50.88               & \textbf{48.61}         \\
SFT + KTO & \textbf{47.00}             & 30.93               & \textbf{22.07}        & \textbf{51.36}      & 48.13                  \\ \bottomrule
\end{tabular}%
}
\caption{Qwen-2.5-Instruct (14B) Results on NL to FOL conversion on the easy, medium and hard splits on the ProverQA dataset. These models are trained on the FOLIO dataset. The ProverQA dataset is only used for evaluation. The results are an average of 6 runs.}
\label{tab:proverqa:qwen}
\end{table}

\subsection{Effect of Context Length}
While we discuss the main results of analyzing the effects of context length on the overall accuracy of the finetuned models, we show the detailed results in Table~\ref{tab:length_analysis}.
\begin{table*}[!ht]
\small
\centering
\resizebox{0.85\textwidth}{!}{
\begin{tabular}{@{}ll|cc|cc|cc@{}}
\toprule
\multirow{2}{*}{\textbf{Model}} & \multirow{2}{*}{\textbf{Setting}} & \multicolumn{2}{c|}{\textbf{Small Context}} & \multicolumn{2}{c|}{\textbf{Medium Context}} & \multicolumn{2}{c}{\textbf{Large Context}} \\ 
& & \textbf{Overall Acc. (↑)} & \textbf{True F1} & \textbf{Overall Acc. (↑)} & \textbf{True F1} & \textbf{Overall Acc. (↑)} & \textbf{True F1} \\
\midrule

\multirow{3}{*}{Llama-3 8B Instruct} & SFT & 59.76 & 62.93 & 55.07 & 59.96 & 19.00 & 21.05  \\
     & SFT + DPO & 65.00 & 68.67 & 48.95 & 55.77 & 21.00 & 17.54\\
     & SFT + KTO & 66.19 & 71.19 & 55.00 & 61.64 & 25.00 & 22.22\\
\midrule
\multirow{3}{*}{Gemma-2 2B Instruct} & SFT & 47.86 & 50.25 & 50.59 & 55.96 & 44.00 & 46.43 \\
     & SFT + DPO & 44.29 & 42.00 & 32.17 & 41.60 & 38.00 & 52.46 \\
     & SFT + KTO & 57.38 & 57.02 & 49.61 & 53.66 & 44.00 & 55.07\\
\midrule
\multirow{3}{*}{Phi-3.5 Mini Instruct} & SFT & 58.10 & 63.77 & 59.08 & 60.90 & 50.00 & 72.73\\
     & SFT + DPO & 58.33 & 60.75 & 61.25 & 59.48 & 71.00 & 77.14\\
     & SFT + KTO & 58.33 & 63.30 & 62.96 & 64.74 & 53.00 & 50.79\\
\bottomrule
\end{tabular}}
\caption{\small{Analysis of the performance of various models across different input context sizes in the validation set of FOLIO. We group the instances in small context (1-2 sentences), medium context (3-5 sentences) and large context (more than 5 sentences). The Overall Acc. column corresponds to the logical correctness metric across the full validation set.
}}
\label{tab:length_analysis}
\end{table*}

\subsection{Semantic Error Evaluation}
\label{sec:app:semanticeval}
Translating natural language problems into logical languages can cause syntactic, logical and semantic errors. While syntactic and logical errors are computable using logical engines such as Prover9, semantic errors are harder to compute, especially when the logical outcomes are correct. 

\paragraph{Automated Evaluation.} For an automated estimate of semantic errors, we convert the premise stories to FOLs and back to textual paragraphs for evaluation. We evaluate the semantic content of generated FOLs by comparing the sentence embeddings of the generated NLs and the input premise stories by cosine similarity. The similarities between the generated NLs and premise paragraphs show the information carried while encoding to and decoding from the FOL representations. We follow different levels of evaluation to make space for jumbled NLs since the premise story need not follow a particular order. Thus, we evaluate the generated NLs against the premise stories in three methods: (i) Firstly, we directly compare the paragraph level embeddings of the premise paragraph and the generated NL. (ii) We allow for jumbling and take the average pair-wise similarity of the two. (iii) Lastly, we allow for jumbling while taking the average of maximum sentence similarity between pairs of sentences between the two. Following this approach, we find that both \textsc{Llama3} and \textsc{GPT3.5} perform similarly with a paragraph average similarity of \textbf{89.13\%} and \textbf{89.26\%} respectively. Further results for each of the evaluation methods can be found in Table ~\ref{tab:NL_conversions}.

\begin{table}[!ht]
\small
\centering
\resizebox{0.48\textwidth}{!}{
\begin{tabular}{@{}l|rr@{}}
\toprule[1pt]
Evaluation metric            & \textsc{LLama3} & \textsc{GPT3.5} \\ \midrule
NL sentence mean similarity  & 54.83  & 55.85  \\
NL sentence max similarity   & 84.09  & 84.67  \\
NL paragraph similarity      & 89.13  & 89.22  \\
FOL sentence mean similarity & 54.83  & 54.57  \\
FOL sentence max similarity  & 84.09  & 79.35  \\
FOL paragraph similarity     & 86.22  & 87.40  \\ \bottomrule
\end{tabular}
}
\caption{Evaluating semantic content of translated FOLs for \textsc{LLama3} and \textsc{GPT3.5} through an autoencoder approach}
\label{tab:NL_conversions}
\end{table}

\end{document}